\documentclass[10pt,twocolumn,letterpaper]{article}

\usepackage{cvpr}              % To produce the CAMERA-READY version
\usepackage[dvipsnames]{xcolor}

\definecolor{cvprblue}{rgb}{0.21,0.49,0.74}
\usepackage[pagebackref,breaklinks,colorlinks,citecolor=cvprblue]{hyperref}
\usepackage{multirow}
\usepackage{pifont}
\usepackage{makecell}
\usepackage{color}
\usepackage{dsfont}

\newcommand{\vv}[1]{\mathbf{#1}}

\usepackage{makecell}
\usepackage{subcaption}
\usepackage{engord}

\title {Towards Robust Event-guided Low-Light Image Enhancement: A Large-Scale Real-World Event-Image Dataset and Novel Approach}

\author{Guoqiang~Liang$^{1}$  \quad Kanghao~Chen$^{1}$ 
\quad Hangyu~Li$^{1}$ \quad Yunfan~Lu$^{1}$
\quad Lin~Wang$^{1,2}$\thanks{Corresponding author}\\
$^{1}$AI Thrust, HKUST(GZ) \quad $^{2}$Dept. of Computer Science and Engineering, HKUST\\
{\tt\small \{gliang041,kchen879,hli886,ylu066\}@connect.hkust-gz.edu.cn, linwang@ust.hk} \\
\small{Project Page: \url{https://vlislab22.github.io/eg-lowlight/}}
}

\begin{document}
\maketitle

\begin{abstract}
Event camera has recently received much attention for low-light image enhancement (LIE) thanks to their distinct advantages, such as high dynamic range.
However, current research is prohibitively restricted by the lack of large-scale, real-world, and spatial-temporally aligned event-image datasets.
To this end, we propose a real-world (indoor and outdoor) dataset comprising over \textbf{30K} pairs of images and events under both low and normal illumination conditions. To achieve this, we utilize a robotic arm that traces a consistent \textbf{non-linear} trajectory to curate the dataset with spatial alignment precision under \textbf{0.03mm}. We then introduce a matching alignment strategy, rendering 90\% of our dataset with errors less than \textbf{0.01s}.
Based on the dataset, we propose a novel event-guided LIE approach, called \textbf{EvLight}, towards robust performance in real-world low-light scenes.
Specifically, we first design the multi-scale holistic fusion branch to extract holistic structural and textural information from both events and images.
To ensure robustness against variations in the regional illumination and noise, we then introduce a Signal-to-Noise-Ratio (SNR)-guided regional feature selection to selectively fuse features of images from regions with high SNR and enhance those with low SNR by extracting regional structure information from events.
Extensive experiments on our dataset and the synthetic SDSD dataset demonstrate
our EvLight significantly surpasses the frame-based methods, \eg,~\cite{cai2023retinexformer} by \textbf{1.14} dB and \textbf{2.62} dB, 
respectively.

\end{abstract}

\begin{figure}[t!]
\centering
    \includegraphics[width=1\linewidth]{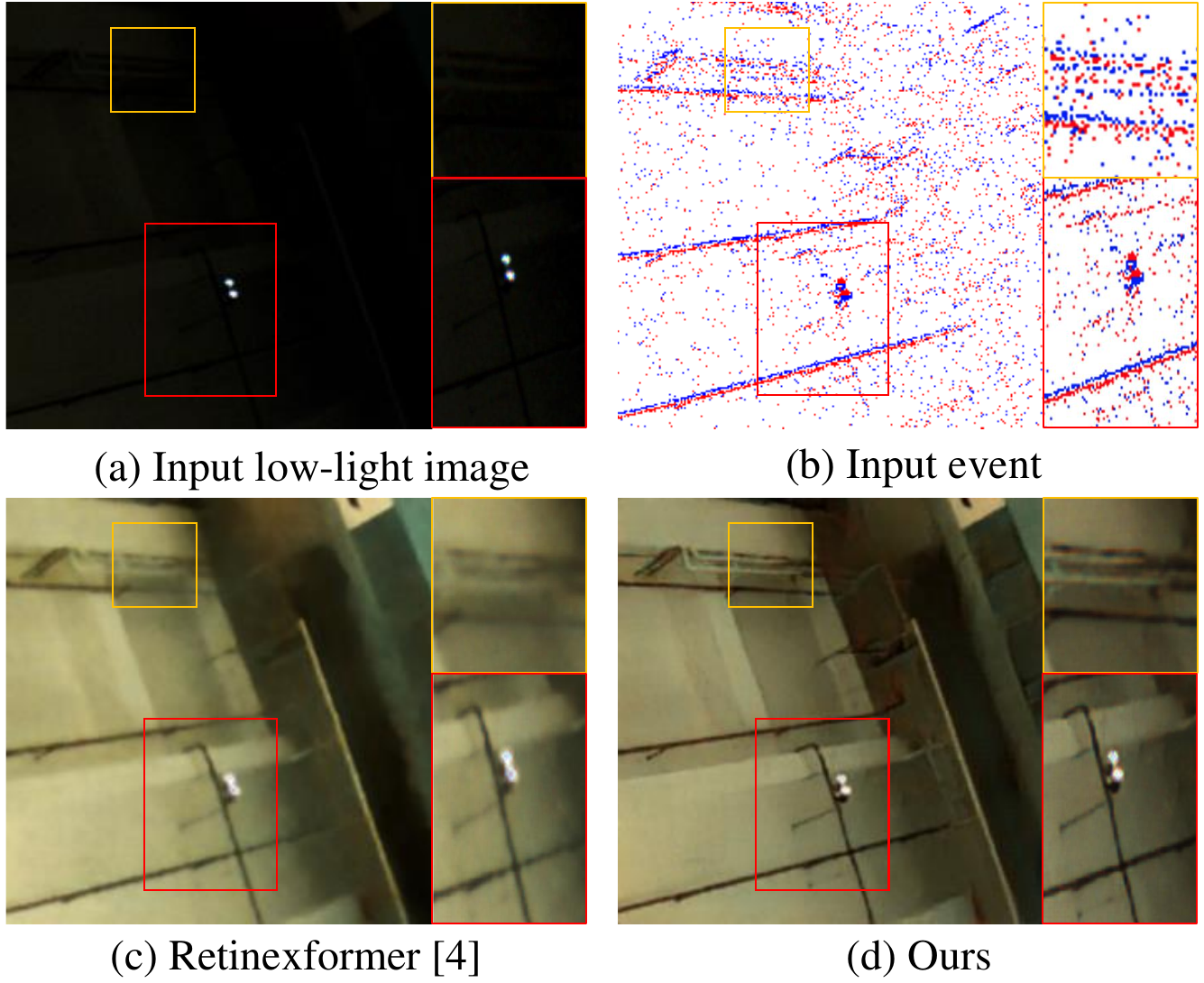}
    \caption{A challenging example of our dataset containing an extremely low-light image (a) and sparse events (b). Compared with the result from a SOTA frame-based method Retinexformer~\cite{cai2023retinexformer} (c), our EvLight (d) not only recovers the structure details (\eg, the pipe on the ceiling) but also avoids over-enhancement and saturation in the bright regions (\eg, the lights).}
    \label{fig:teaserfigure}
\centering
\end{figure}

\section{Introduction}
Images captured under sub-optimal lighting conditions often exhibit various types of degradation such as poor visibility, noise, and inaccurate color~\cite{li2021low}.
For this reason, low-light image enhancement (LIE) serves as an essential task in ameliorating low-light image quality. LIE is crucial for downstream tasks,~\eg, face detection~\cite{ma2022toward,yu2021single} and nighttime semantic segmentation~\cite{pan2023towards}.
Recently, with the emergence of deep learning, abundant frame-based methods have been proposed, ranging from enhancing contrast~\cite{zhang2021beyond12}, removing noise~\cite{xu2020learning} to correcting color~\cite{wang2019underexposed}. Although the performance has been remarkably boosted, these methods often suffer from unbalanced exposure and color distortion when the visual details, \eg, edges, provided by frame-based cameras are less distinctive, as shown in Fig.~\ref{fig:teaserfigure} (c).

\begin{figure*}[t]
\centering
    \includegraphics[width=1\linewidth]{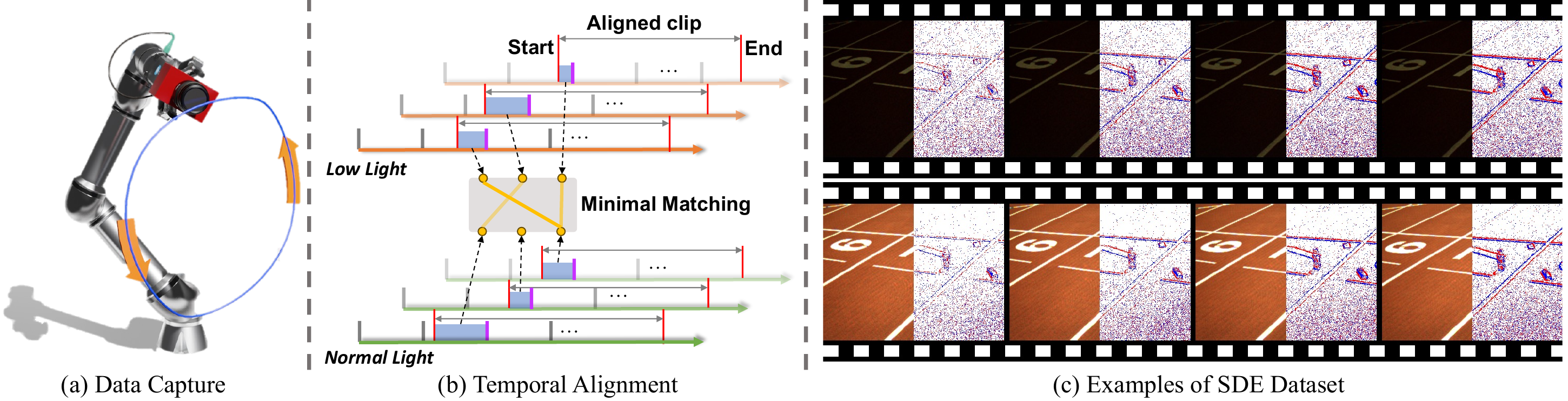}
    \caption{(a) An illustration of collecting spatially-aligned image-event dataset by mounting a DAVIS 346 event camera on the robotic arm and recording the sequences with the same trajectory receptively. (b) An overview of our matching alignment strategy. (c) An example of our dataset with images and paired events captured in low-light (with an ND8 filter) and normal-light conditions.}
    \label{fig:dataset-temporal-alignment}
\centering
\end{figure*}

Event cameras are bio-inspired sensors that generate event streams with high dynamic range (HDR), high temporal resolution, \etc~\cite{scheerlinck2019ced,zheng2023deep}. 
However, few research efforts have been made in combining both frame-based and event cameras to address the LIE task~\cite{zhang2020learning,liu2023low,jiang2023event,liang2023coherent} to date.  
A hurdle is the prohibitive lack of large-scale real-world datasets with spatial-temporally aligned images and events.
For example,~\cite{zhang2020learning} proposes an unsupervised framework without the need for paired event-image data, and~\cite{liu2023low,liang2023coherent} leverage the synthetic datasets for training.
Nonetheless, these methods are less competent for applications in real-world low-light scenarios.
LIE dataset~\cite{jiang2023event} is a real-world event-image dataset with paired low-light/normal-light sequences, obtained by simply adjusting indoor lamplight (artificial light fluctuations) and outdoor exposure time while maintaining a fixed camera position. 
Thus, similar to the previous frame-based dataset SMID~\cite{chen2019seeing22}, this dataset is only limited to static scenes.

In this paper, we propose a large-scale real-world dataset, named \textbf{SDE} dataset -- containing over 30K pairs of spatio-temporally aligned images and events (see examples in Fig.~\ref{fig:dataset-temporal-alignment} (c)) -- captured under both low-light and normal-light conditions (Sec.~\ref{sec:dataset}).
To construct such a dataset, the inherent difficulty stems from the complexities involved in ensuring precise spatial and temporal alignment between paired low-light and normal-light sequences, especially for dynamic scenes in nonlinear motion.
To achieve this, we design a robotic alignment system to guarantee spatial alignment, where a DAVIS346 event camera~\cite{taverni2018front} is mounted on a Universal UR5 robotic arm, see Fig.~\ref{fig:dataset-temporal-alignment} (a).
Our system shows a remarkable spatial accuracy with an error margin of merely 0.03mm, a significant improvement over the frame-based dataset, SDSD~\cite{wang2021seeing21} with the error of 1mm. 
Moreover, unlike the setup of uniform linear motion in SDSD and the static scene in the LIE dataset~\cite{jiang2023event}, our system embraces non-linear motions with complex trajectories. 
This significantly enhances the diversity of our dataset for real-world scenarios.
As for temporal alignment, a direct way to obtain aligned sequences is to clip them according to the specific motion start and end timestamps.
However, even with the same camera and robot setting, the intervals (\textcolor{blue}{blue regions} in Fig.~\ref{fig:dataset-temporal-alignment} (b)) between motion start timestamps (\textcolor{red}{left red line}) and
the timestamps of the initial frame (\textcolor{magenta}{magenta line}) in each clipped sequence are different, resulting in random temporal errors.
To this end, we propose a novel matching alignment strategy to reduce the temporal discrepancies.

Buttressed by the dataset, we propose an event-guided LIE approach, called \textbf{EvLight}, towards the robust performance in real-world low-light scenes. 
The underlying premise is that -- while low-light images deliver crucial color contents and events offer essential edge details -- 
both modalities may be corrupted by different kinds of noise, yielding different noise distributions. Therefore, directly fusing the features of both modalities, as commonly done in~\cite{jiang2023event}, may also aggravate the noise in different regions of the two inputs, as shown in the blue box area in Fig.~\ref{fig:visual-our-indoor} (g).

To tackle these problems, our key idea is to fuse event and image features holistically, followed by a selective region-wise manner to extract the textural and structural information with the guidance of Signal-to-Noise-Ratio (SNR) prior information. 
To ensure robustness against variations in the regional illumination and noise, we further introduce an SNR-guided feature selection to extract features of images from regions with high SNR and those of events from regions with low SNR.
This preserves the regional textural and structural information (Sec.~\ref{sec:regional}).
Then, we design an attention-based holistic fusion branch to coarsely extract holistic structural and textural information from both events and images (Sec.~\ref{sec:holistic}).
Finally, a fusion block with channel attention is employed to fuse the holistic feature with the regional feature of images and events.

We conduct extensive experiments by comparing with the frame-based~\eg,~\cite{cai2023retinexformer} and event-guided~\eg,~\cite{liu2023low} methods on our real-world dataset and SDSD dataset (frame-based dataset)~\cite{wang2021seeing21} with events generated from the event simulator~\cite{hu2021v2e}.
The experiments show that our EvLight works decently for enhancing diverse underexposed images under extremely low-light conditions, as depicted in Fig.~\ref{fig:teaserfigure}.

\section{Related Work}
\noindent\textbf{LIE Datasets. }
The performance of learning-based methods heavily relies on the quality of the training datasets~\cite{fu2023dancing} for either images~\cite{wei2018deep,chen2018learning,bychkovsky2011learning} or videos~\cite{lee2023humanpose, wang2019ehsc, chen2019seeing22,wang2021seeing21,fu2023dancing,jiang2019smoid}. For example, SDSD \cite{wang2021seeing21} obtains a pair of videos under various light conditions from a scene by mounting the camera on a mechatronic system.
In this paper, we mainly focus on the event-image datasets. 
A summary of existing image-event datasets for low-light enhancement is shown in Tab.~\ref{tab:dataset_comparison}.
EvLowLight~\cite{liang2023coherent} only includes low-light images/events without corresponding normal-light images/events as ground truth, while DVS-Dark~\cite{zhang2020learning} provides unpaired low-light/normal-light images/events.
LIE~\cite{jiang2023event} is a real-world image-event dataset, captured by adjusting the camera's light intake in a static scene, wherein events are triggered by the light changes (indoor) and exposure times (outdoor). 
In contrast, we present a real-world dataset with over 30$K$ spatially and temporally aligned image-event pairs (both indoor and outdoor), using a robotic alignment system, considering the non-linear motion.

\begin{table}[!t]
    \centering
    \resizebox{1\linewidth}{!}{
\begin{tabular}{cccccc}
\hline
Dataset     & Release                 & Dynamic Scene          & With Ground Truth         & Numbers        \\ \hline
DVS-Dark \cite{zhang2020learning} & \ding{55} & \ding{51} & \ding{55} & 17,765  \\
 LIE \cite{jiang2023event}            & \ding{55}   & \ding{55}        & \ding{51}     & 2,231          \\
  EvLowLight \cite{liang2023coherent}            & \ding{55}   & \ding{51}        & \ding{55}     & ---          \\
                           Ours           &\ding{51}                     & \ding{51}    & \ding{51}   & 31,477          \\ \hline
\end{tabular}
}
\caption{A summary of existing real-world image-event datasets. Note that images in DVS-Dark are gray-scale.}
    \label{tab:dataset_comparison}
\end{table}

\noindent\textbf{Frame-based LIE.}
Frame-based methods for low-light image enhancement can be divided into non-learning-based methods~\cite{arici2009histogram1,nakai2013color2,guo2016lime,fu2016weighted9,xu2020star10} and learning-based methods~\cite{wei2018deep,zhang2019kindling13,zhang2021beyond12,wu2022uretinex5,fu2023you6,cai2023retinexformer,wang2019underexposed,xu2022snr4,xu2023low7,wang2023low8,wu2023learning}.
Non-learning-based methods typically rely on handcrafted features, such as histogram equalization~\cite{arici2009histogram1,nakai2013color2} and the Retinex theory~\cite{guo2016lime,fu2016weighted9,xu2020star10}. 
Nonetheless, these methods lead to the absence of adaptivity and efficiency~\cite{wu2022uretinex5}.
With the development of deep learning, an increasing number of learning-based methods have emerged, which can be bifurcated as Retinex-based methods~\cite{wei2018deep,zhang2019kindling13,zhang2021beyond12,wu2022uretinex5,fu2023you6,cai2023retinexformer} and non-Retinex-based methods~\cite{wang2019underexposed,xu2022snr4,xu2023low7,wang2023low8,wu2023learning}.
Specially, SNR-Aware~\cite{xu2022snr4} collectively exploits Signal-to-Noise-Ratio-aware transformers and convolutional models to dynamically enhance pixels with spatial-varying operations.
However, these frame-based approaches often result in blurry outcomes and low Structural Similarity (SSIM) due to the buried edge in low-light images.

\noindent\textbf{Event-based LIE.}
Event cameras enjoy HDR and provide rich edge information even under low-light scenes~\cite{zheng2023deep}.
Zhang \etal~\cite{zhang2020learning} focuses on reconstructing grayscale images from low-light events but faces challenges in preserving original details using only brightness changes from events.
Recently, some researchers have utilized events as guidance for low-light image enhancement~\cite{jiang2023event, jin2023event}, low-light video enhancement~\cite{liu2023low,liang2023coherent}, and deblurring for low-light images~\cite{zhou2021delieve}.
ELIE~\cite{jiang2023event} utilizes a residual fusion module to blend event and image for low light enhancement.
Liu \textit{et al.}~\cite{liu2023low} address artifacts in prior low-light video enhancement methods by synthesizing events from adjacent images for intensity and motion information, and propose a fusion transform module to fuse these event features with image features.
EvLowLight~\cite{liang2023coherent} establishes temporal coherence by jointly estimating motion from both events and frames while ensuring the spatial coherence between events and frames with different spatial resolutions.
However, these methods directly fuse features extracted from events and images without considering the discrepancy of the noise at the different local regions in events and images.

\section{Our SDE Dataset}
\label{sec:dataset}

Capturing paired dynamic sequences from real-world scenes presents a formidable challenge, primarily attributed to the complexity involved in ensuring spatial and temporal alignment under varying illumination conditions.
The first line of approaches employs a stereo camera system to simultaneously record the identical scenes, using non-linear transformations and cropping like DPED~\cite{ignatov2017dslr}.
However, it struggles with SIFT keypoint computation and matching~\cite{lowe2004distinctive} in the low light. This hinders the identification of overlapped video segments.
The second line of approaches~\cite{jiang2019smoid,lee2023humanpose} constructs an optical system incorporating a beam splitter, allowing two cameras to capture a unified view. 
Nonetheless, achieving impeccable alignment with such systems remains challenging, resulting in spatial misalignments, as mentioned in \cite{lee2023humanpose,rim2020real,liang2023coherent}.
The third line of approaches, \eg, SDSD~\cite{wang2021seeing21} proposes a mechatronic system mounting the camera on an electric slide rail to capture low-light/normal-light videos separately (two rounds).
However, SDSD is constrained by the limited \textit{linear} motion of the electric slide rail.
Differently, we design a robotic alignment system, equipped with an event camera to capture paired RGB images and events, under both low-light and normal-light conditions. 
Our system features the non-linear motions with complex trajectories.

\begin{figure*}[t]
\centering
    \includegraphics[width=.98\linewidth]{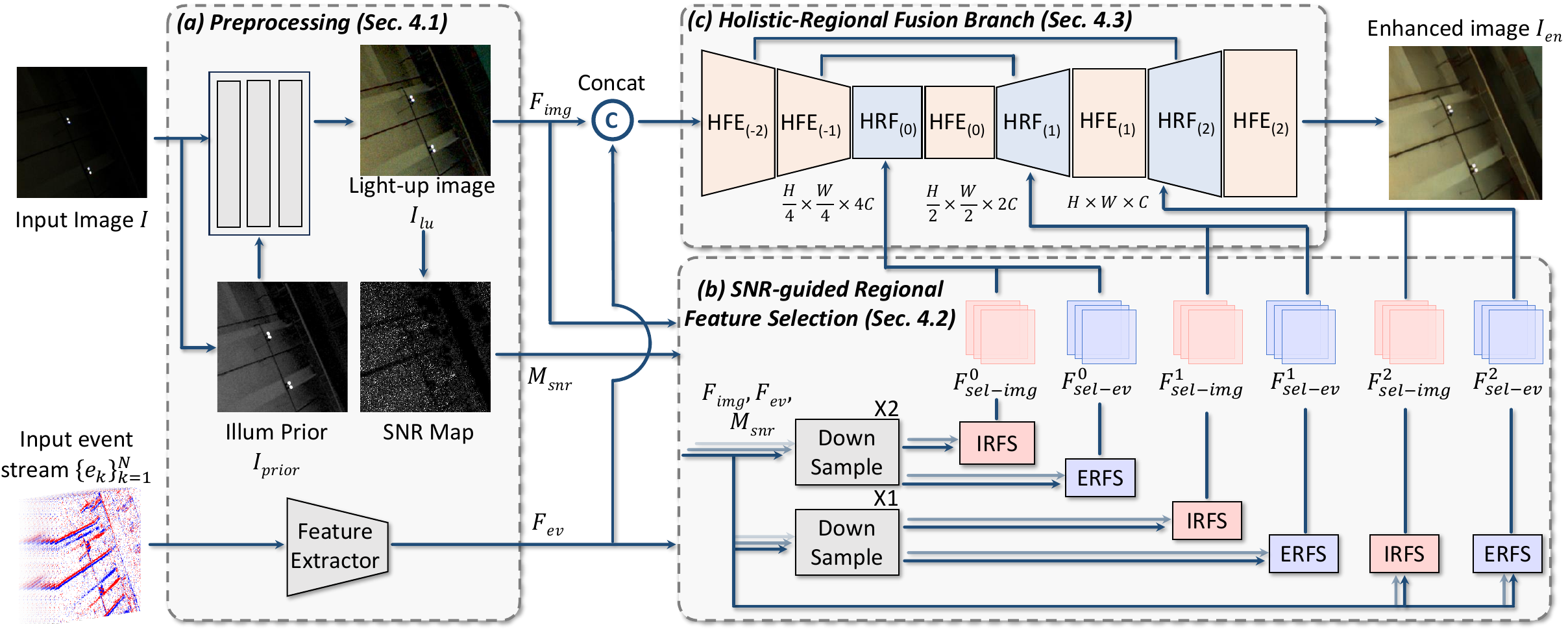}
    \caption{\small \textbf{An overview of our framework}. Our method consists of three parts, \textbf{(a)} Preprocessing (Sec.~\ref{sec:preprocessing}), \textbf{(b)} SNR-guided Regional Feature Selection (Sec.~\ref{sec:regional}), and \textbf{(c)} Holistic-Regional Fusion Branch (Sec.~\ref{sec:holistic}). Specifically, SNR-guided Regional Feature Selection consists of two parts: Image-Regional Feature Selection (IRFS) and Event-Regional Feature Selection (ERFS). Additionally, Holistic-Regional Fusion Branch encompasses Holistic Feature Extraction (HFE) and Holistic-Regional Feature Fusion (HRF).}
    \label{fig:framework}
\centering
\end{figure*}

\noindent\textbf{1) Data Capture with Spatial Alignment.}
To ensure the spatial alignment of paired sequences, a robotic arm (Universal UR5), exhibiting a minimal repeated error of 0.03mm, is equipped to capture sequences following an identical trajectory. 
We set the robotic system with a pre-defined trajectory and a DAVIS 346 event camera with fixed parameters, \eg exposure time.
Firstly, paired image and event sequences are acquired under normal lighting conditions.
Subsequently, an ND8 filter is applied to the camera lens, which facilitates the capture of low-light sequences while maintaining consistent camera parameters, such as exposure time and frame intervals.

\noindent\textbf{2) Temporal Alignment of Low-light/Normal-light sequences.}
The alignment of SDSD~\cite{wang2021seeing21} dataset involves a manual selection of the initial and final frames of each paired video, based on the motion states depicted in the videos, leading to inevitable bias.
To mitigate this problem, initial temporal alignment is performed by trimming the sequences based on the start and end timestamps of a pre-defined trajectory.
However, even with consistent settings for exposure time and frame intervals, there exists a variable time interval between the start timestamp of the trajectory and the first frame timestamp captured post-initiation of the trajectory in each sequence.
The bias causes the misalignment between each low-light image and its normal-light image pair, particularly in complex motion paths.

To achieve further alignment, we introduce a matching alignment strategy, wherein sequences from each scene are captured multiple times to minimize the alignment error to the largest extent, as shown in Fig.~\ref{fig:dataset-temporal-alignment} (b).
Practically, we capture 6 paired event-image sequences per scene —three in low-light and three in normal-light conditionals. 
These 6 sequences are trimmed to the predefined trajectory's start and end timestamps, ensuring uniform content across all videos.
Subsequently, the time intervals between the trajectory's start timestamps and the initial frame timestamps of each trimmed sequence are calculated.
As shown in Fig.~\ref{fig:dataset-temporal-alignment} (b), the time intervals (\textcolor{blue}{blue regions}) of 6 sequences are different, and we match the low-light sequence with the normal-light sequence, which has the minimal absolute errors of their time intervals; thus, we can reduce the misalignment caused by the random time interval.
With the matching alignment strategy, we achieve a remarkable precision, with 90\% of the datasets, exhibiting temporal alignment errors below 0.01s, and maximum errors of 0.013s and 0.027s for our indoor and outdoor datasets, respectively.

\section{The Proposed EvLight Framework}
\label{sec:method}
Based on our SDE dataset, we further propose a novel event-guided LIE framework, called \textbf{EvLight}, as 
 depicted in Fig.~\ref{fig:framework}.
 % summary challenge and goals 
Our goal is to selectively fuse the features of the image and events to achieve robust performance for event-guided LIE.
EvLight takes the low-light image $\vv{I}$ and paired event stream $\left\{\vv{e}_k \right\}_{k=1}^N$ with $N$ events as inputs and outputs an ehnanced image $\vv{I}_{en}$.
Our pipeline consists of three components:
\textbf{1)} Preprocessing,
\textbf{2)} SNR-guided Regional Feature Selection, and
\textbf{3)} Holistic-Regional Fusion Branch.

\noindent \textbf{Event Representation.}
Given an event stream $\left\{\vv{e}_k \right\}_{k=1}^N$, we follow \cite{rebecq2019events} to obtain the event voxel grid $\vv{E}$ by assigning the polarity of each event to the two closest voxels. The bin size is set to 32 in all the experiments.

\subsection{Preprocessing}
\label{sec:preprocessing}
\noindent \textbf{Initial Light-up.}
As demonstrated in recent frame-based LIE methods~\cite{cai2023retinexformer,wang2023low8,xu2023low7}, coarsely enhancing the low-light image benefits the image restoration process and further boosts the performance.
For simplicity, we follow Retinexformer~\cite{cai2023retinexformer} for the initial enhancement.
As shown in the Fig.~\ref{fig:framework}, we estimate the initial light-up image $\vv{I}_{lu}$ as:
\begin{equation}
    \vv{I}_{lu} = \vv{I} \odot \vv{L}, \\ \vv{L} = \mathcal{F}(\vv{I}, \vv{I}_{prior}),
\end{equation}
where $\vv{I}_{prior}=max_c(\vv{I})$ denotes the illumination prior map, with $max_c$ denoting the operation that computes the max values for each pixel across channels.
$\mathcal{F}$ outputs the estimated illumination map $\vv{L}$, which is then applied to the input image $\vv{I}$ through a pixel-wise dot product. 

\noindent \textbf{The SNR Map.}
Following the previous approaches~\cite{buades2005non,dabov2006image,xu2022snr4}, we estimate the SNR map based on the initial light-up image $\vv{I}_{lu}$ and make it an effective prior for the SNR-guided regional feature selection in Sec.~\ref{sec:regional}.
Given the initial light-up image $\vv{I}_{lu}$, we first convert it into grayscale one $\vv{I}_{g}$, \ie, ${\vv{I}}_{g}\in \mathbb{R}^{H\times W}$, followed by computing the SNR map $\vv{M}_{snr} = \tilde{\vv{I}}_g / abs(\vv{I}_{g} - \tilde{\vv{I}}_g)$, where $\tilde{\vv{I}}_g$ is the denoised counterpart of $\vv{I}_{g}$. 
In practice, similar to SNR-Net~\cite{xu2022snr4}, the denoised counterpart is calculated with the mean filter.

\noindent \textbf{Feature Extraction.} 
Image feature $\vv{F}_{img}$ of light-up image $\vv{I}_{lu}$ and event feature $\vv{F}_{ev}$ of the event voxel grid $\vv{E}$ are initially extracted with $conv3\times3$. 

\subsection{SNR-guided Regional Feature Selection}
\label{sec:regional}

\begin{figure}[t]
\centering
    \includegraphics[width=0.95\linewidth]{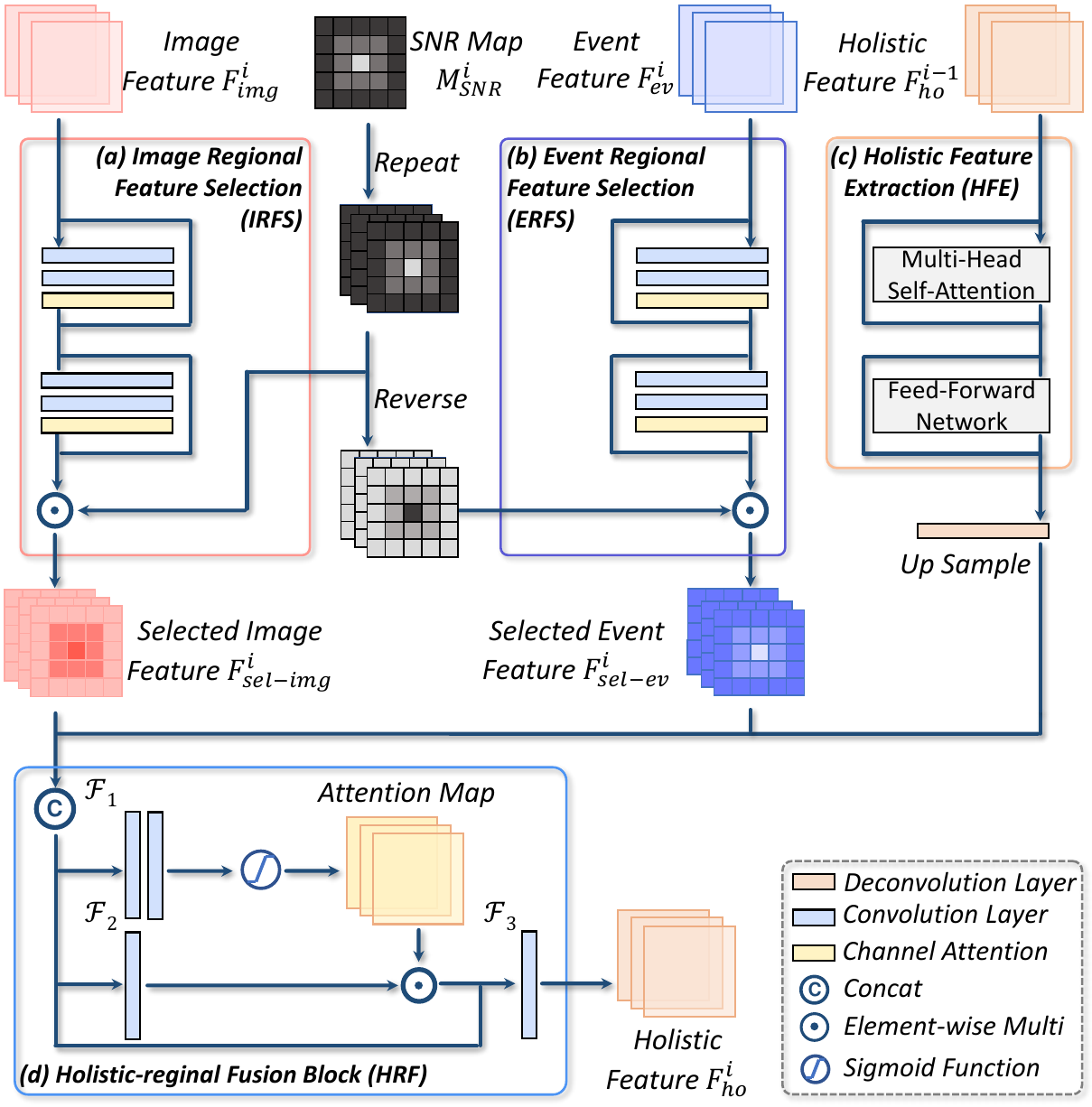}
    \caption{Details of each block in SNR-guided Regional Feature Selection and Holistic-Regional Fusion Branch's decoder.}
    \label{fig:snr_block}
\centering
\end{figure}
In this section, we aim to \textit{selectively extract the regional features from either images or events}.
We design an image-regional feature selection (IRFS) block to select image feature with higher SNR values, thereby obtaining image-regional feature, less affected by noise.
However, SNR map assigns low SNR values to not only high-noise regions but also edge-rich regions.
Consequently, solely extracting features from regions with high SNR values can inadvertently suppress edge-rich regions.
To address this, we introduce an event-regional feature selection (ERFS) block for enhancing edges in areas with poor visibility and high noise.

As shown in Fig.~\ref{fig:framework}, inputs of this module include the image feature $\vv{F}_{img}$, the event feature $\vv{F}_{ev}$, and the SNR map $\vv{M}_{snr}$.
Firstly, the image feature $\vv{F}_{img}$ and event feature $\vv{F}_{ev}$ are down-sampled with $conv4\times4$ layers with the stride of 2 and SNR map $\vv{M}_{snr}$ undergoes an averaging pooling with the kernel size of 2.
These donwsampling operations are represented as `\textit{Down Sample}' in Fig.~\ref{fig:framework} and we attain different scale image feature $\vv{F}^i_{img} \in \mathbb{R}^{\frac{H}{2^{2-i}} \times \frac{W}{2^{2-i}} \times 2^{2-i}C}$, event feature $\vv{F}^i_{ev} \in \mathbb{R}^{\frac{H}{2^{2-i}} \times \frac{W}{2^{2-i}} \times 2^{2-i}C}$, and SNR map $\vv{M}^{i}_{snr} \in \mathbb{R}^{\frac{H}{2^{2-i}} \times \frac{W}{2^{2-i}}}$  where $i = 0,1,2$.
Then, the image feature $\vv{F}^i_{img}$ and event feature $\vv{F}^i_{ev}$ are selected with the guidance of SNR map $\vv{M}^i_{snr}$ in IRFS block, and ERFS block. 
These two blocks then output the selected image features $\vv{F}^{i}_{sel-img}$ and event features $\vv{F}^{i}_{sel-ev}$, respectively. 
We now describe the details of these two blocks.

\noindent \textbf{Image-Regional Feature Selection (IRFS) Block.} 
As depicted in Fig.~\ref{fig:snr_block} (a), for an image feature $\vv{F}^i_{img}$, we initially process it through two residual blocks~\cite{he2016deep} to extract regional information and yield the output $\hat{\vv{F}}^i_{img}$.
Each block comprises two $conv3\times3$ layers and an efficient channel attention layer~\cite{wang2020eca}. 
The SNR map $\vv{M}^i_{snr}$ is then expanded  
along the channel
to align with the image feature's channel dimensions.
Then, we normalize it and make it within the range of $[0, 1]$. 
We then apply a predefined threshold on the SNR map to attain $\hat{\vv{M}}^{i}_{snr}$.
To emphasize regions with higher SNR values and attain the selected image feature $\vv{F}^{i}_{sel-img}$, we perform an element-wise multiplication $\odot$  between the extended SNR map and the image feature $\hat{\vv{F}}^i_{img}$, formulated as:
\begin{equation}
    \begin{aligned}
    \vv{F}^{i}_{sel-img} &= \hat{\vv{M}}^{i}_{snr} \odot \hat{\vv{F}}^i_{img}. \\
    \end{aligned}
\end{equation}

\noindent \textbf{Event-Regional Feature Selection (ERFS) Block.}
Edge-rich regions in the initial light-up image, particularly those underexposed, exhibit low SNR values.
Additionally, we observe that events in high SNR regions (\eg, well-illuminated smooth planes) are predominantly leak noise and shot noise events. 
Consequently, we design the ERFS block that utilizes the inverse of the SNR map to selectively enhance edges in low-visibility, high-noise areas, and to suppress noise events in sufficiently illuminated regions.
The initial processing in this block follows a similar architecture to that used for the IRFS block, with $\vv{F}^i_{ev}$ as the input and $\hat{\vv{F}}^i_{ev}$ as the output.
Given the SNR map $\hat{\vv{M}}^{i}_{snr}$, we obtain the reserve of SNR map $\vv{\bar{M}}^{i}_{snr}$ by  $\mathds{1}$ - $\hat{\vv{M}}^{i}_{snr}$. To obtain the selected event-regional feature $\vv{F}^{i}_{sel-ev}$, the element-wise multiplication product $\odot$ between the reserve of SNR map and the event feature is carried out, which is formulated as:
\begin{equation}
    \begin{aligned}
    \vv{F}^{i}_{sel-ev} &= \vv{\bar{M}}^{i}_{snr} \odot \hat{\vv{F}}^i_{ev}. \\
    \end{aligned}
\end{equation}

\subsection{Holistic-Regional Fusion Branch}
\label{sec:holistic}
In this section, we aim to \textit{extract the holistic features from both the event features and image features, so as to build up long-range channel-wise dependencies between them.
}
Besides, the holistic features are enhanced with the selected image-regional and event-regional features in the holistic-region feature fusion process.

\begin{table*}[t]
    \centering
    \setlength{\tabcolsep}{3pt}
    \resizebox{1.0\textwidth}{!}{
\begin{tabular}{cccccccccccccc}
\hline
\multirow{2}{*}{Input} & \multirow{2}{*}{Method} & \multicolumn{3}{c}{SDE-in}  & \multicolumn{3}{c}{SDE-out}   & \multicolumn{3}{c}{SDSD-in}    & \multicolumn{3}{c}{SDSD-out}\\
& & \multicolumn{1}{l}{PSNR$\uparrow$}  & \multicolumn{1}{l}{PSNR*$\uparrow$} & \multicolumn{1}{l}{SSIM$\uparrow$}  & \multicolumn{1}{l}{PSNR$\uparrow$} & \multicolumn{1}{l}{PSNR*$\uparrow$} & \multicolumn{1}{l}{SSIM$\uparrow$}  & \multicolumn{1}{l}{PSNR$\uparrow$} & \multicolumn{1}{l}{PSNR*$\uparrow$} & \multicolumn{1}{l}{SSIM$\uparrow$}  & \multicolumn{1}{l}{PSNR$\uparrow$} & \multicolumn{1}{l}{PSNR*$\uparrow$} & \multicolumn{1}{l}{SSIM$\uparrow$}  \\ \hline
Event Only                   & E2VID+ (ECCV'20) \cite{reducingsimtoreal}           
&15.19              &15.92              &0.5891                                &15.01              &16.02                    &0.5765                            &13.48                    &13.67              &0.6494                      
&16.58      &17.27       &0.6036     \\ \hline
\multirow{4}{*}{Image Only}  & SNR-Net (CVPR'22) \cite{xu2022snr4}          &20.05         &21.89        &0.6302                       &22.18         &22.93 &0.6611                      &24.74         &25.30 &0.8301                            &24.82         &26.44 &0.7401                 \\
                            & Uformer (CVPR'22) \cite{wang2022uformer}          &21.09         &22.75       &\underline{0.7524}                      &22.32         &23.57 &\underline{0.7469}                      &24.03         &25.59 &\underline{0.8999}                            &24.08         &25.89 &\underline{0.8184}                 \\
                             & LLFlow-L-SKF (CVPR'23)~\cite{wu2023learning}     &20.92         &22.22 &0.6610                        &21.68         &23.41 &0.6467                       &23.39              &24.13 &0.8180                              &20.39         &24.73 &0.6338               \\
                             & Retinexformer (ICCV'23)~\cite{cai2023retinexformer}    &21.30        &23.78  &0.6920                       &\underline{22.92}         &23.71 &0.6834                       &25.90         &25.97 &0.8515                             &\underline{26.08}         &\underline{28.48} &0.8150                 \\ \hline
\multirow{3}{*}{Image+Event} & ELIE (TMM'23)~\cite{jiang2023event}              &19.98         &21.44 &0.6168                          &20.69         &23.12 &0.6533                      &{27.46}         &{28.30} &{0.8793}                             &23.29              &28.26 &0.7423          \\
                             & eSL-Net (ECCV'20)~\cite{wang2020event}         &21.25       &23.19 &0.7277                        &22.42        &\underline{24.39} &0.7187                              &24.99         &25.72 &0.8786                         &24.49          &26.36 &0.8031             \\
                             & Liu \etal (AAAI'23)~\cite{liu2023low}         &\underline{21.79}       &\underline{23.88} &0.7051                            &22.35        &23.89 &0.6895                              &\underline{27.58}         &\underline{28.43} &{0.8879}                              &23.51          &27.63 &0.7263                 \\
                             & Ours                       &\textbf{22.44}          &\textbf{24.81} &\textbf{0.7697}                           &\textbf{23.21}        &\textbf{25.60} &\textbf{0.7505}                     &\textbf{28.52}       &\textbf{29.73} &\textbf{0.9125}                        &\textbf{26.67}        &\textbf{30.30} &\textbf{0.8356}             \\ \hline
\multicolumn{1}{l}{}         & \multicolumn{1}{l}{}    & \multicolumn{1}{l}{}     & \multicolumn{1}{l}{}     & \multicolumn{1}{l}{}      & \multicolumn{1}{l}{}     & \multicolumn{1}{l}{}     & \multicolumn{1}{l}{}      & \multicolumn{1}{l}{}     & \multicolumn{1}{l}{}                          
\end{tabular}
    }
    \caption{\textbf{Comparisons on our SDE dataset and SDSD~\cite{wang2021seeing21} dataset}. The highest result is highlighted in \textbf{bold} while the second highest result is highlighted in \underline{underline}. Since E2VID+~\cite{reducingsimtoreal} can only reconstruct grayscale images, its metrics are calculated in grayscale.}
    \label{tab:main_result}
\end{table*}

Fig.~\ref{fig:framework} (c) depicts our holistic-regional fusion branch, which employs a UNet-like architecture~\cite{ronneberger2015u} with the skip connections.
This branch takes the concatenated feature of image $\vv{F}_{img}$ and event $\vv{F}_{ev}$ from the preprocessing stage (Sec.~\ref{sec:preprocessing}) as the input and the enhanced image $\vv{I}_{en}$ as the output.
In the contracting path, there are 2 layers and the output of each layer is ${\vv{F}}^{i+1}_{ho} \in \mathbb{R}^{\frac{H}{2^{2-|i+1|}} \times \frac{W}{2^{2-|i+1|}} \times 2^{2-|i+1|}C}$ where $i = -2,-1$.
In the $i$-th layer, the holistic feature $\vv{F}^i_{ho}$ first undergoes the holistic feature extraction (HFE) block.
Then with a strided $conv4\times4$ down-sampling operation, the holistic feature $\vv{F}^{i+1}_{ho}$ is obtained.
In the expansive path, the output of each layer is ${\vv{F}}^i_{ho}$ where $i = 0, 1, 2$. 
As shown in Fig.~\ref{fig:snr_block}, the holistic feature ${\vv{F}}^{i-1}_{ho}$ is processed with the HFE block and $\hat{\vv{F}}^{i-1}_{ho}$ is produced.
Then, the holistic feature $\hat{\vv{F}}^{i-1}_{ho}$ is up-sampled with a strided $deconv2\times2$ and it is fused with the selected regional image $\vv{F}^{i}_{sel-img}$ and event features $\vv{F}^{i}_{sel-ev}$ in the holistic-regional fusion (HRF) block.

\noindent \textbf{Holistic Feature Extraction (HFE) Block.} 
As shown in Fig.~\ref{fig:snr_block} (c), holistic feature extraction is mainly composed of a multi-head self-attention module and a feed-forward network.
Given a holistic feature ${\vv{F}}^{i-1}_{ho}$, the feature can be processed as:
\begin{equation}
    \begin{aligned}
       \hat{\vv{F}}^{i-1}_{mid} &= \text{Attention}({\vv{F}}^{i-1}_{ho})+{\vv{F}}^{i-1}_{ho}, \\
       \hat{\vv{F}}^{i-1}_{ho} &= 
       \text{FFN}(\text{LN}(\hat{\vv{F}}^{i-1}_{mid})) + 
       \hat{\vv{F}}^{i-1}_{mid},
    \end{aligned}
\end{equation}
where $\hat{\vv{F}}^{i-1}_{mid}$ is the middle output, LN is the layer normalization, FFN represents the feed-forward network, and Attention signifies the channel-wise self-attention, analogous to the multi-head attention mechanism employed in~\cite{zamir2022restormer}.

\noindent \textbf{Holistic-Regional Fusion (HRF) Block.} 
This block first concatenates the selected image features $\vv{F}^i_{sel-img}$, selected event features $\vv{F}^i_{sel-ev}$, and up-sampled holistic features $\hat{\vv{F}}^{i-1}_{ho}$.
This concatenated feature $\vv{F}^i_{cat}$ is then passed through $conv3\times3$ layers to generate a spatial attention map.
Sequentially, the element-wise multiplication is performed between the attention map and the concatenated features, which can be denoted as:
\begin{equation}
    \begin{aligned}
       {\vv{F}}^i_{ho} &= \mathcal{F}_3(\sigma(\mathcal{F}_1(\vv{F}^i_{cat})) \odot \mathcal{F}_2(\vv{F}^i_{cat}) + \vv{F}^i_{cat}),
    \end{aligned}
\end{equation}
where $\mathcal{F}_i$ is the convolution operation indicated in Fig.~\ref{fig:snr_block} (d). $\sigma$ and $\odot$ denote the Sigmoid function and the element-wise production, respectively.

\noindent \textbf{Optimization.} The loss function $\mathcal{L}$ utilized for training is articulated as: $\mathcal{L} = \sqrt{||\vv{I}_{en}-\vv{I}_{gt}||^2 +\epsilon^2} + \lambda||\Phi({\vv{I}}_{en}) - \Phi({\vv{I}}_{gt}) ||_1 $, 
where $\lambda$ is a hyper-parameter, $\epsilon$ is set to $10^{-4}$, $\vv{I}_{en}$ and $\vv{I}_{gt}$ denote the enhanced and ground truth images, and $\Phi$ represents feature extraction using the Alex network~\cite{krizhevsky2012imagenet}.

\section{Experiments}
\noindent\textbf{Implementation Details:}
We employ the Adam optimizer~\cite{kingma2014adam} for all experiments, with learning rates of $1e-4$ and $2e-4$ for SDE and SDSD datasets, respectively. 
Our framework is trained for 80 epochs with a batch size of 8 using an NVIDIA A30 GPU. 
We apply random cropping, horizontal flipping, and rotation for data augmentation. 
The cropping size is 256 $\times$ 256, and the rotation angles include 90, 180, and 270 degrees.

\noindent\textbf{Evaluation Metrics:} We use the peak-signal-to-noise ratio (PSNR)~\cite{hore2010image} and SSIM~\cite{wang2004image} for evaluation.
Following the finetuning of the overall brightness of predicted results in previous methods~\cite{zhang2019kindling13,wu2023learning}, we introduce the PSNR* as the metric to assess image restoration effectiveness beyond light fitting.
The calculation of PSNR* is formulated as:
\begin{equation}
    \begin{aligned}
    \vspace{-5pt}
    \text{PSNR*} &= \text{PSNR}(\vv{I}_{en}\times{\vv{R}}_{gt-en},\vv{I}_{gt}), \\
    {\vv{R}}_{gt-en} &= \text{Mean}(\text{Gray}(\vv{I}_{gt})) / \text{Mean}(\text{Gray}(\vv{I}_{en})),
    \end{aligned}
\end{equation}
where $\vv{I}_{en}$, $\vv{I}_{gt}$, Gray, Mean, and PSNR represent the enhanced image, the ground-truth image, the operation of converting RGB images to grayscale ones, the operation of getting mean value, and the operation of calculating PSNR value, respectively.

\begin{figure*}[t]
\centering
    \includegraphics[width=0.98\linewidth]{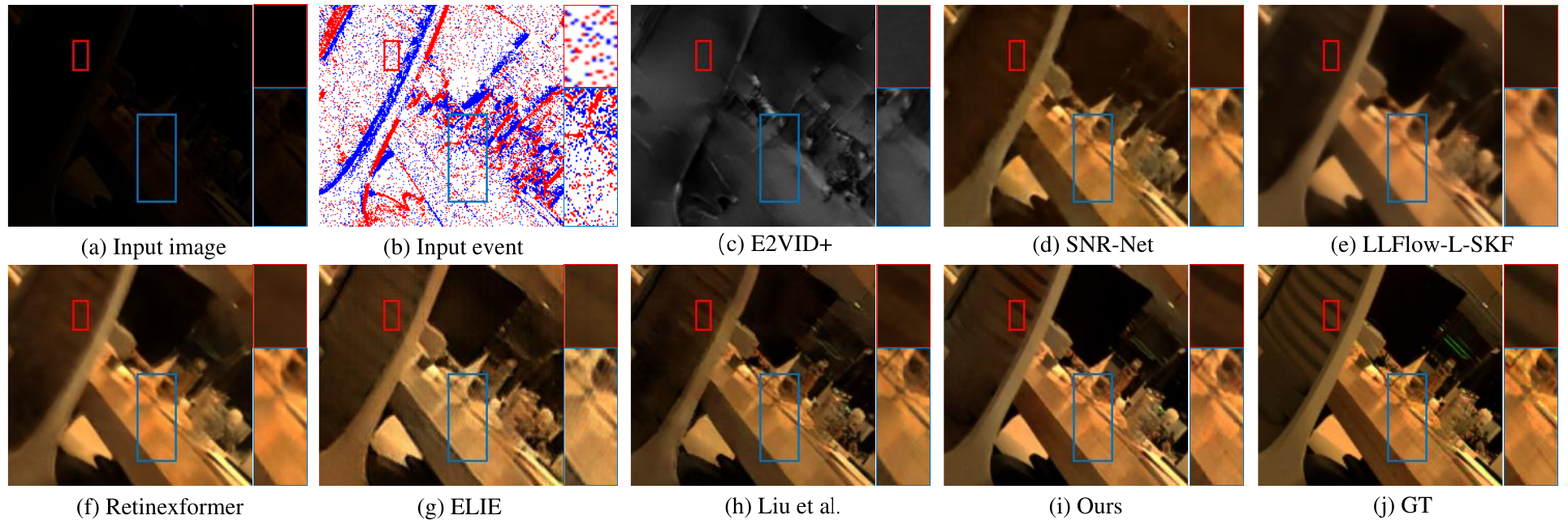}
    \caption{Qualitative results on our SDE-in dataset.}
    \label{fig:visual-our-indoor}
\centering
\end{figure*}

\begin{figure*}[t]
\centering
    \includegraphics[width=0.98\linewidth]{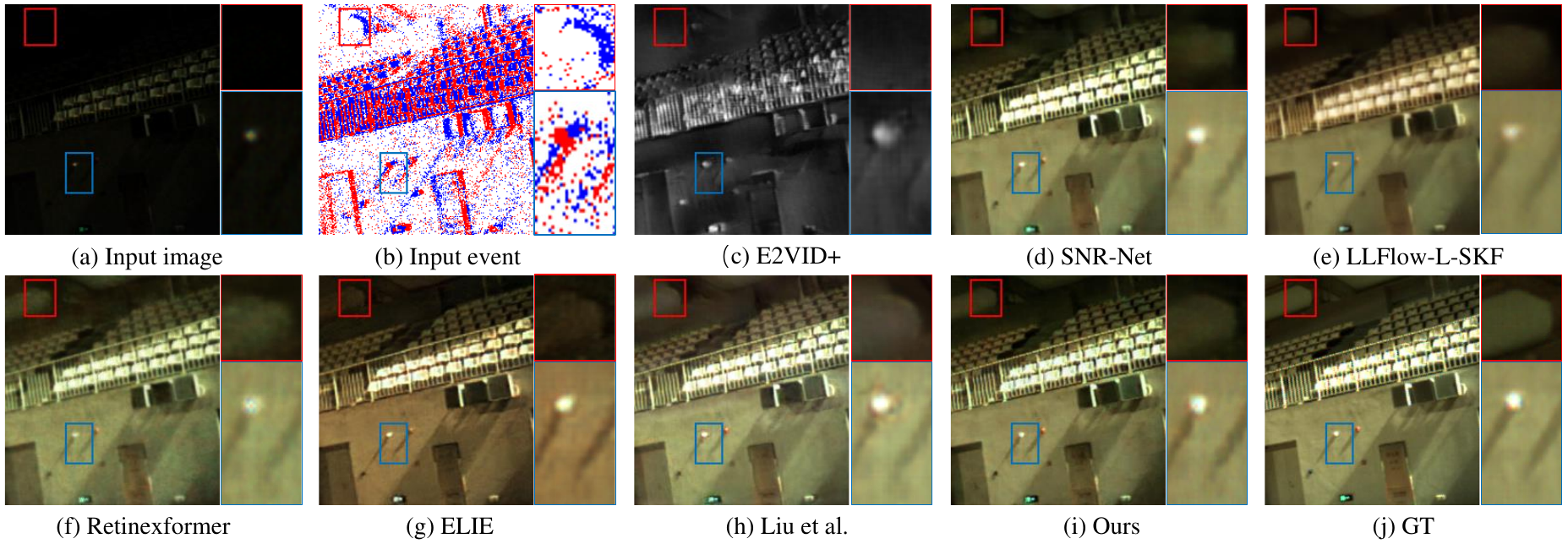}
    \caption{Qualitative results on our SDE-out dataset.}
    \label{fig:visual-our-outdoor}
\centering
\end{figure*}
\noindent\textbf{Datasets:}
\textbf{1) SED dataset~}
contains 91 image+event paired sequences (43 indoor and 48 outdoor sequences) captured with a DAVIS346 event camera~\cite{scheerlinck2019ced} which outputs RGB images and events with the resolution of $346\times260$.
For all collected sequences, 76 sequences are selected for training, and 15 sequences are for testing. 
\noindent\textbf{2) SDSD dataset~\cite{wang2021seeing21}~}
provides paired low-light/normal-light videos with $1920\times1080$ resolution containing static and dynamic versions.
We choose the dynamic version for simulating events and employ the same dataset split scheme as in SDSD~\cite{wang2021seeing21}: 125 paired sequences for training and 25 paired sequences for testing.
We first downsample the original videos to the same resolution ($346\times260$) of the DAVIS346 event camera. 
Then, we input the resized images to the event simulator v2e~\cite{hu2021v2e} to synthesize event streams with noise under the default noisy model.

\begin{figure*}[t]
\centering
    \includegraphics[width=1\linewidth]{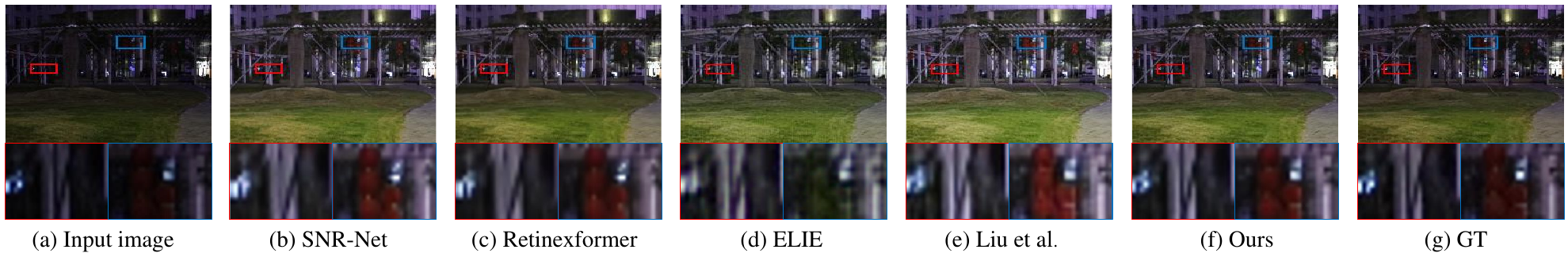}
    \caption{Qualitative results on SDSD dataset~\cite{wang2021seeing21}.}
    \label{fig:visual-sdsd-outdoor}
\centering
\end{figure*}

\subsection{Comparison and Evaluation}
We compare our method with recent methods with three different settings: 
\textbf{(I)} the experiment with events as input, including E2VID+~\cite{reducingsimtoreal}. \textbf{(II)} the experiment with a RGB image as input, including SNR-Net~\citep{xu2022snr4}, Uformer~\cite{wang2022uformer}, LLFlow-L-SKF~\cite{wu2023learning}, and Retinexformer~\cite{cai2023retinexformer}.
\textbf{(III)} the experiment with a RGB image and paired events as inputs, including ELIE~\citep{jiang2023event}, eSL-Net~\cite{wang2020event}, and Liu \etal~\cite{liu2023low}.
We reproduced ELIE~\cite{jiang2023event} and Liu \etal~\cite{liu2023low} according to the descriptions in the papers, while the others are retrained with the released code.
We replace the event synthesis module in~\cite{liu2023low} by inputting events captured with the event camera or generated from the event simulator~\cite{hu2021v2e}.

\noindent \textbf{Comparison on our SDE Dataset:}
Quantitative results in Tab.~\ref{tab:main_result} showcase our method's superior performance on the SDE dataset, outperforming baselines with higher PSNR by 0.65 dB for SDE-in and 0.29 dB for SDE-out.
To assess image restoration effectiveness beyond light fitting, we computed PSNR* and our method also notably surpasses SOTA techniques, achieving higher PSNR* by 0.93 dB for SDE-in and 1.21 dB for SDE-out.
This marks a significant validation of our approach for low-light image enhancement.
Qualitatively, as depicted in Fig.~\ref{fig:visual-our-indoor} and Fig.~\ref{fig:visual-our-outdoor} for indoor and outdoor scenes respectively, our method effectively reconstructs clear edges in dark areas (\eg, the red box areas in Fig.~\ref{fig:visual-our-indoor} and Fig.~\ref{fig:visual-our-outdoor}), surpassing frame-based methods like Retinexformer~\cite{cai2023retinexformer} and event-guided approaches such as Liu \etal~\cite{liu2023low}.
Moreover, our method demonstrates less color distortion and noise on challenging regions (e.g., the wall in Fig.~\ref{fig:visual-our-outdoor}) than LLFlow-L-SKF~\cite{wu2023learning} and ELIE~\cite{jiang2023event}, and Retinexformer~\cite{cai2023retinexformer}, underscoring our method's robustness.

\noindent\textbf{Comparison on the SDSD Dataset:}
To evaluate our method's generalization, we conducted comparisons on the SDSD dataset~\cite{wang2021seeing21}, with quantitative outcomes detailed in Tab.~\ref{tab:main_result}. 
Our method outperforms baselines significantly in PSNR, PSNR*, and SSIM, leading by more than 0.94 dB for SDSD-in and 0.59 dB for SDSD-out. 
Although ELIE and Liu~\etal~\cite{liu2023low} surpass frame-based methods in SDSD-in dataset, they suffer from the overfitting in SDSD-out dataset which is demonstrated by the substantial disparity between PSNR and PSNR*.
Qualitatively, as shown in Fig.~\ref{fig:visual-sdsd-outdoor}, our method effectively restores underexposed images to more detailed structures, as highlighted in the red box area.
Moreover, ELIE~\cite{jiang2023event} tends to produce color distortions, as visible in the blue box area of Fig.~\ref{fig:visual-sdsd-outdoor} (d).

\subsection{Ablation Studies and Analysis}
We conduct ablation studies on SDE-in dataset to assess the effectiveness of each module of our method.
The basic implementation, without SNR-guided regional feature selection as described in Sec.~\ref{sec:regional}, is called the \textit{Base} model.

\begin{figure}[t]
\centering
    \includegraphics[width=0.90\linewidth]{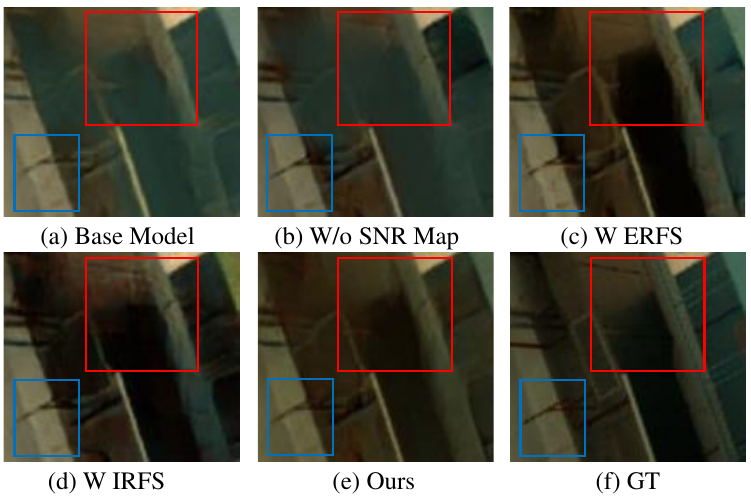}
    \caption{Visualization of ablation results.}
    \label{fig:ablation}
\centering
\end{figure}

\begin{table}
\centering
\setlength{\tabcolsep}{5pt}
\resizebox{0.45\textwidth}{!}{
\begin{tabular}{c|cccc}
\hline
                      & \makecell[c]{Regional Feature Selection} & SNR-guided        & PSNR        & SSIM\\ 
\hline 
1             &\ding{55}  &\ding{55}       &21.58  &0.7001\\
2             &\ding{51}  &\ding{55}        &21.86  &0.7490\\
3             &\ding{51}  &\ding{51}   &\textbf{22.44} &\textbf{0.7697}\\
\hline
\end{tabular}
}
\caption{Ablation of SNR-guided regional feature selection.
\label{tab:ablation_regioanl}}
\end{table}

\begin{table}
\centering
\setlength{\tabcolsep}{18pt}
\resizebox{0.45\textwidth}{!}{
\begin{tabular}{c|cccc}
\hline
 &\makecell[c]{IRFS} & \makecell[c]{ERFS} & PSNR & SSIM\\ 
\hline
1 &\ding{55}             &\ding{55}         &21.58  &0.7001 \\
2 &\ding{51}             &\ding{55}         &21.92  &0.7108\\
3 &\ding{55}             &\ding{51}         &22.18  &0.7525\\
4 &\ding{51}             &\ding{51}          &\textbf{22.44}  &\textbf{0.7697}\\
\hline
\end{tabular}
}
\caption{Impact of each module of SNR-guided regional feature selection.\label{tab:ablation_snr}}
\end{table}

\noindent\textbf{Impact of Events:}
To reveal the impact of events, we conduct experiments on the \textit{Base} model. 
The variant excluding events attains a PSNR of 21.35 dB and an SSIM of 0.6985, whereas adding events results in a 0.23 dB improvement in PSNR and a 0.002 improvement in SSIM.
However, the \textit{Base} model cannot fully explore the potential of events demonstrated by the limited improvement in SSIM.

\noindent\textbf{Impact of SNR-guided regional feature selection:}
To verify it, we conduct an ablation study in Tab.~\ref{tab:ablation_regioanl}.
We replace the SNR map with an all-ones matrix and remove the whole selection module (the \textit{Base} model).
Compared with the \textit{Base} model (\engordnumber{1} row), regional feature selection with an all-ones matrix (\engordnumber{2} row) and SNR-guided regional feature selection (\engordnumber{3} row) yield 0.28 dB and 0.86 dB increase in PSNR, respectively, demonstrating the necessity of regional features and the SNR map.
Although regional feature selection with an all-ones matrix and \textit{Base} model both have color distortion (\eg, the red box in Fig.~\ref{fig:ablation} (a), (b)), (b) has better structure details than (a).

\noindent\textbf{Impact of IRFS and ERFS:}
To verify them, we conduct an ablation study in Tab.~\ref{tab:ablation_snr}.
Compared with the \textit{Base} model (\engordnumber{1} row), image-regional feature selection (IRFS, \engordnumber{2} row), event-regional feature selection (ERFS, \engordnumber{3} row), and the combination of them (\engordnumber{4} row) yields the 0.34 dB, 0.60 dB, and 0.86 dB increase in PSNR, respectively, demonstrating the necessity of the IRFS and ERFS block.
As shown in Fig.~\ref{fig:ablation}, IRFS (d) or ERFS (c) can reduce the color distortion that appears in the \textit{Base} model (a).
With both IRFS and ERFS blocks, our results deliver the best visual quality (\eg, red box and blue box in Fig.~\ref{fig:ablation}).

\noindent\textbf{Generalization Ability:}
To assess the generalization capability of our EvLight, we carry out an experiment on the CED~\cite{scheerlinck2019ced} and MVSEC~\cite{zhu2018multivehicle} with the model trained on our SDE dataset.
Moreover, we use the model, trained on the synthetic events from the SDSD dataset~\cite{wang2021seeing21} to evaluate the generalization capacity on real events of our SDE dataset. 
\textit{Detailed visual results are available in Suppl. Mat. }

\section{Conclusion}
This paper presented a large-scale real-world event-image dataset, called SDE, curated via a non-linear robotic path for high-fidelity spatial and temporal alignment, encompassing low and normal illumination conditions.
Based on the real-world dataset, we designed a framework, EvLight, towards robust event-guided low-light image enhancement, which adaptively fuse the event and image features in a holistic and region-wised manner resulting in robust and superior performance.

\noindent\textbf{Limitations and Future Work:}
Due to inherent limitations of DAVIS346 event cameras, RGB images in our SDE dataset may exhibit partial chromatic aberrations and the moiré pattern.
In the future, we will improve our hardware system to enable synchronous triggering of robots and event cameras, thereby significantly reducing labor costs associated with repetitive collection.

\section{Acknowledgment}
This paper is supported by the National Natural Science Foundation of China (NSF) under Grant No. NSFC22FYT45 and the Guangzhou City, University and Enterprise Joint Fund under Grant No.SL2022A03J01278.

{
    \small
    \bibliographystyle{ieeenat_fullname}
    \bibliography{ref}
}

\end{document}